\pdfoutput=1

\documentclass[11pt]{article}

\usepackage{acl}

\usepackage{times}
\usepackage{latexsym}
\usepackage{multirow}
\usepackage{overpic}

\usepackage{times}
\usepackage{latexsym}
\usepackage{arydshln}

\usepackage[T1]{fontenc}
\usepackage{booktabs}

\usepackage[utf8]{inputenc}

\usepackage{microtype}

\newcommand{\intermediatesize}{\fontsize{9.8}{11.4}\selectfont}

\usepackage{inconsolata}

\title{LLMs in Biomedical: A Study on Named Entity Recognition}

\author{Masoud Monajatipoor \\
  UCLA \\
  \texttt{\intermediatesize{monajati@ucla.edu}} \\\And
  Jiaxin Yang \\
  UCLA \\
  \texttt{\intermediatesize{yangjiaxin0821@ucla.edu }} \\\And
  Joel Stremmel \\
  Optum \\
  \texttt{\intermediatesize{joel\_stremmel@optum.com }} \\
  \AND
  Melika Emami \\
  Optum \\
  \texttt{\intermediatesize{melika.emami@optum.com }} \\
  \\\And
  Fazlolah Mohaghegh \\
  Optum \\
  \texttt{\intermediatesize{ehsan.mohaghegh@optum.com  }} \\
  \\\And
  Mozhdeh Rouhsedaghat \\
  USC \\
  \texttt{\intermediatesize{rouhseda@usc.edu}} \\
  \\\And
  Kai-Wei Chang  \\
  UCLA \\
  \texttt{\intermediatesize{kwchang@cs.ucla.edu}} \\
  }

\begin{document}
\maketitle
\begin{abstract}


Large Language Models (LLMs) demonstrate remarkable versatility in various NLP tasks but encounter distinct challenges in biomedical due to the complexities of language and data scarcity. This paper investigates LLMs application in the biomedical domain by exploring strategies to enhance their performance for the NER task.
Our study reveals the importance of meticulously designed prompts in the biomedical. Strategic selection of in-context examples yields a marked improvement, offering $\sim 15-20\%$ increase in F1 score across all benchmark datasets for biomedical few-shot NER. 
Additionally, our results indicate that integrating external biomedical knowledge via prompting strategies can enhance the proficiency of general-purpose LLMs to meet the specialized needs of biomedical NER.
Leveraging a medical knowledge base, our proposed method, DiRAG, inspired by Retrieval-Augmented Generation (RAG), can boost the zero-shot F1 score of LLMs for biomedical NER. Code is released at \url{https://github.com/masoud-monajati/LLM_Bio_NER}
\end{abstract}

\section{Introduction} 
LLMs such as GPT4 have demonstrated exceptional capabilities across diverse tasks and domains \cite{espejel2023gpt, dai2023can, dong2022survey}. 
These models could have a revolutionary impact on healthcare; however, their integration into medical research and practice has been slow \cite{zhou2023universalner, vaishya2023chatgpt, nori2023capabilities} and it is crucial to examine the unique challenges presented by the biomedical field that contribute to this discrepancy.  
Specifically, LLMs encounter challenges in medical Information Extraction \cite{gutierrez2022thinking, moradi2021gpt} due to the scarcity of high-quality biomedical data in their pretraining, and the need for a nuanced comprehension of the text for this task \cite{gu2023distilling}. 
Medical entities can have multiple synonyms and abbreviations, complicating their recognition by models \cite{grossman2021deep}.
Furthermore, context sensitivity is even more critical in the biomedical compared to the general domain. 
The specificity of entity types and the complexity of their interrelations
necessitate a level of background knowledge that standard prompts may fail to provide. LLMs are primarily exposed to vast amounts of generic text data limiting their effectiveness in managing the intricate nuances of medical language \cite{kumari2023large,karabacak2023embracing}.

In this paper, we concentrate on NER, a foundational task for various applications  
such as recruiting patients for clinical trials, searching biomedical literature, or building models that predict the progression of disease based on free-text notes.

In our initial analysis, we broaden the scope of TANL \cite{paolini2021structured} and DICE \cite{ma2022dice}, two text-to-text formats initially proposed for model training, adapting their use to prompt design specifically for biomedical NER. Our findings reveal that the relative effectiveness of the resulting prompt pattern varies based on specific dataset characteristics. 
Subsequently, we investigate the importance of example selection via In-Context Learning (ICL) and demonstrate the value of nearest neighbor example selection using pre-trained biomedical text encoders when performing biomedical NER.
A key question that arises in the deployment of LLMs concerns the comparative advantage of closed-source LLMs versus open-source ones. In our third study, we shed light on this question by presenting an assessment of performance and cost across various experiments. 
Furthermore, we explore the integration of external medical knowledge to refine LLM capabilities  \cite{gao2023retrieval, zakka2024almanac}. Leveraging the insights gained from these techniques, we present a novel data augmentation method incorporating a medical knowledge base, e.g.,  UMLS \cite{bodenreider2004unified}, which substantially improves zero-shot biomedical NER. 

\section{Background and Preliminaries}

\paragraph{Prompt engineering}
Prompt tuning \cite{white2023prompt, lester2021power, ding2021openprompt} as its own research field shows that skillfully crafted prompts can significantly enhance LLM understanding 
for complex tasks \cite{lu2021fantastically, kaddour2023challenges, webson2021prompt}. 
Researchers have explored different prompt formats for IE tasks with LLMs \cite{wang2023instructuie, gutierrez2022thinking, wang2023gpt} including more work around knowledge insertion for prompt augmentation \cite{seo2024retrieval, chen2023minprompt} 
Another type of prompting is ICL \cite{brown2020language}, where LLMs use a limited set of "input-output" pairs within the prompt along with a query input as demonstrations of what the task output should be. 
In this realm, \citet{liu2021makes, min2022rethinking, gao2023makes} demonstrated that choosing targeted in-context examples over random sampling leads to more accurate model responses. 

\paragraph{Named Entity Recognition}
GPT-NER \cite{wang2023gpt} was one of the first methods to incorporate a unique symbol to transform the sequence tagging task into text generation via ICL with GPT-3 \cite{brown2020language}, achieving performance on par with fully supervised baselines. Following this work, \citet{gutierrez2022thinking, moradi2021gpt} showed that LLMs are not skilled few-shot learners in the biomedical domain. 
However, recent advancements, such as GPT-4, have increased LLM performance on many tasks \cite{tian2024opportunities, hu2024zero, nori2023capabilities} including in the biomedical domain \cite{hu2024improving}.
In the direction of knowledge distillation from LLMs \cite{wang2023instructuie, gu2023distilling}, \citet{zhou2023universalner} presented UniNER, a targeted distillation technique coupled with instruction tuning to develop an efficient open-domain NER model. Our research draws from these works and explores the capabilities of LLMs for biomedical NER, employing prompt design, strategic ICL example selection, and data augmentation via an external knowledge base to enhance performance.

\paragraph{Problem definition}
Assume data samples are represented as $(X,Y)$ and the goal is to develop a model, denoted as $f: (X \times T) \rightarrow Y$, where X signifies the input set, T represents a predetermined set of entity types, and Y denotes the set of entity types. The task is to predict the entity type of each input word among the set T. We followed the standard practice of using the F1 score for evaluation purposes in both mention/token-level analyses.

\paragraph{Datasets}
We used three biomedical NER datasets with different entity types: I2B2 \cite{uzuner20112010} which includes test, treatment, and problem entities, NCBI-disease \cite{dougan2014ncbi} consisting of the disease entity, and BC2GM \cite{smith2008overview} containing the gene entity.  

\section{Influence of Input-Output Format}
\begin{figure}[ht]
\centering
\begin{overpic}
[width=0.45\textwidth]{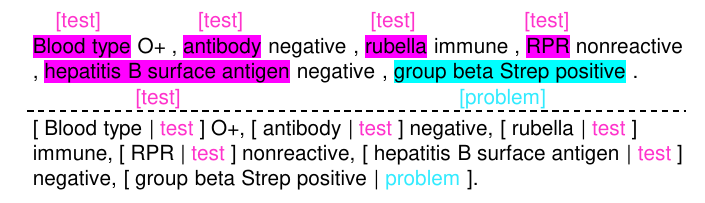}
    \put(0,16){\rotatebox{90}{\scriptsize{{input}}}}
    \put(0,1){\rotatebox{90}{\scriptsize{{output}}}}
\end{overpic}
\caption{TANL input/output format for NER task.} \label{tanl}
\end{figure}

Recent studies demonstrated the importance of prompt engineering for various tasks \cite{wang2023chatcad, gao2023leveraging, nori2023can}. 
We studied 
the influence of input-output format by adapting TANL \cite{paolini2021structured} and DICE \cite{ma2022dice} 
for biomedical NER. In TANL, the task is framed as a translation task which involves augmenting the text by tagging entity types for each word directly within the text.
The method is exemplified in Fig \ref{tanl}, showcasing how the text incorporates entity types.

\begin{figure}[ht]
\centering
\begin{overpic}
[width=0.48\textwidth]{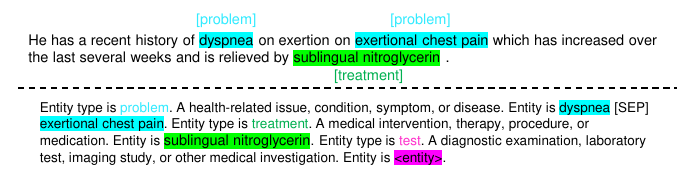}
    \put(0,16){\rotatebox{90}{\scriptsize{{input}}}}
    \put(0,1){\rotatebox{90}{\scriptsize{{output}}}}
\end{overpic}
\caption{DICE input/output format for NER task.} \label{dice}
\end{figure}

Then, the generated output is decoded into the BIO format \cite{ramshaw1999text} for the assessment. In the refined DICE format, the input-output format involves adding a description for each entity type in a template following DEGREE \cite{hsu2021degree}. Given an input text and corresponding labels, the desired output should be the input followed by the phrase "entity type is <entity\_type>. <entity\_description>. entity is <entity>" for each class label, e.g., \textit{test}, \textit{treatment}, and \textit{problem} in the I2B2 dataset. Then, we expect the model to output the same template filling out the <entity> with the corresponding entities in the given text as demonstrated in Fig \ref{dice}. For the entity types with no matched entities in the sentence, the output returns <entity> token in the output. Examples for the NCBI-disease and BC2GM datasets are presented in Appendix \ref{sec:appendix}.

Our experiments in Table \ref{dicevsTANL} reveal that neither format consistently outperforms the other; rather, the effectiveness of each format varies depending on the complexity of the dataset and model size.
To maintain consistency in the rest of our experiments, we opted for the TANL format, in which the input-output relationship exhibits a more straightforward pattern.

\begin{table}
\small
\begin{center}
  \centering
  \resizebox{\linewidth}{!}{
  \begin{tabular}{ccccc}
    \toprule
    Model & input-output format & I2B2 & NCBI-disease &  BC2GM\\
    & & M/T & M/T &  M/T\\
    \midrule
    \multirow{2}{*}{GPT-3.5-turbo}
    & DICE  & 41.2 /50.0 & 45.3 /\textbf{62.0} & \textbf{43.3} /\textbf{55.6}  \\
    & TANL  & \textbf{52.9}/\textbf{59.7} & \textbf{46.5}/51.3 & 39.1/50.8 \\
    
    \midrule
    \multirow{2}{*}{GPT-4} 
    & DICE  & 58.8/70.1 & \textbf{68.1}/\textbf{77.8} &  \textbf{57.1}/67.9 \\
    & TANL  & \textbf{61.9}/\textbf{73.5} &  67.5/70.0 & 56.4/\textbf{69.6}  \\
    
    \bottomrule
  \end{tabular}
  }
  \caption{TANL vs. DICE format with GPT-3.5-turbo/GPT-4 . The superiority of any single format varies with the complexity of the dataset and model size.} 
  \label{dicevsTANL}
  \end{center}
\vspace{-1pt}
\end{table}

\section{In-Context Examples Selection: A Key to Improving ICL Outcomes}
In-context examples can be randomly chosen from the training set; however, researchers have demonstrated that the performance of ICL depends on the order and similarity of ICL examples to the test samples \cite{liu2021makes,min2022rethinking, gao2023makes}. \citet{liu2021makes} presented Knn-Augmented in-conText Example selection (KATE). KATE identifies in-context examples selectively using nearest neighbor search on example embeddings, leading to better performance than random example selection.
We tested KATE on TANL formatted examples with 16-shot ICL using four different LM encoders (w/o fine-tuning) to produce example embeddings. We used  
MPNET \cite{song2020mpnet} for its popularity and performance on sentence embedding benchmarks \cite{reimers-2019-sentence-bert}
, SimCSE \cite{gao2021simcse} for its documented performance as an alternative to standard sentence transformers, and BioClinicalBERT \cite{alsentzer2019publicly} and BioClinicalRoBERTa \cite{domains} for 
their dominance on clinical data tasks \cite{lehman2023we}. 

Our results summarized in Table \ref{KATE} show that strategic in-context example selection via KATE outperforms random selection. 
BioClinicalRoBERTa achieved the best results among all example encoders tested.  The strong performance of BioClinicalBERT and BioClinicalRoBERTa underscores the importance of using LM encoders pretrained on biomedical text when applying KATE for biomedical NER. 

\begin{table}[ht]
\small
\begin{center}
  \centering
  \resizebox{\columnwidth}{!}{
  \begin{tabular}{ccccc}
    \toprule
    Model & KATE vs RS & I2B2 & NCBI-disease &  BC2GM \\
    & & M/T & M/T &  M/T\\
    \midrule
    \multirow{6}{*}{GPT-3.5-turbo (ICL)} 
    & RS & 52.9/59.7 & 46.6/51.3 & 39.1/50.8 \\
    \cdashline{2-5}
    & BioClinicalRoBERTa & 66.1/77.4 & \textbf{68.0}/77.7 & \textbf{61.6}/\textbf{72.5} \\
    & BioClinicalBERT  & \textbf{67.0}/\textbf{78.9}  & 67.6/\textbf{78.8} & 60.9/72.0  \\
    \cdashline{2-5}
    & MPNET  & 65.3/76.7 & 63.7/76.7 &  59.1/70.0 \\
    & SimCSE  & 65.2/76.1  & 61.6/76.1 & 57.8/68.8  \\
    \cline{2-5}
    & \cite{hu2024improving} & 49.3/ - & - & - \\
    \midrule
    \multirow{6}{*}{GPT4 (ICL)} 
    & RS & 67.7/73.5 & 62.6/70.0 & 59.2/69.6 \\
    \cdashline{2-5}
    & BioClinicalRoBERTa & 81.2/\textbf{88.4} & \textbf{79.3}/\textbf{88.3} & \textbf{72.4}/\textbf{80.7} \\
    & BioClinicalBERT  & \textbf{81.7}/88.1  & \textbf{79.3}/88.0 & 71.9/79.4  \\
    \cdashline{2-5}
    & MPNET & 80.7/87.5 & 79.8/87.4 &  71.1/80.2 \\
    & SimCSE  & 79.6/86.6  & 77.3/86.5 & 69.9/77.9  \\

    \cline{2-5}
    & \cite{hu2024improving} & 59.3/ - & - & - \\
    
    \midrule
    BioBERT & fully supervised & - /87.3 & - /\textbf{89.1} &  - /83.8 \\
    BioClinicBERT & fully supervised & - /87.7 & - /89.0 &  - /81.7 \\
    BioClinicRoBERTa & fully supervised & - /\textbf{89.7} & - /89.0 &  - /\textbf{87.0} \\
    
    \bottomrule
  \end{tabular}
  }
  \caption{16-shot ICL for Random example selection (RS) vs. KATE method Vs MLMs with Mention/Token-level (M/T) analysis. KATE significantly outperforms random sampling in all settings, and LMs pre-trained on biomedical text outperform general domain encoders.}
  \label{KATE}
  \end{center}
\vspace{-1pt}
\end{table}

\begin{figure*}[ht]
\centering
\begin{overpic}
[width=0.85\textwidth]{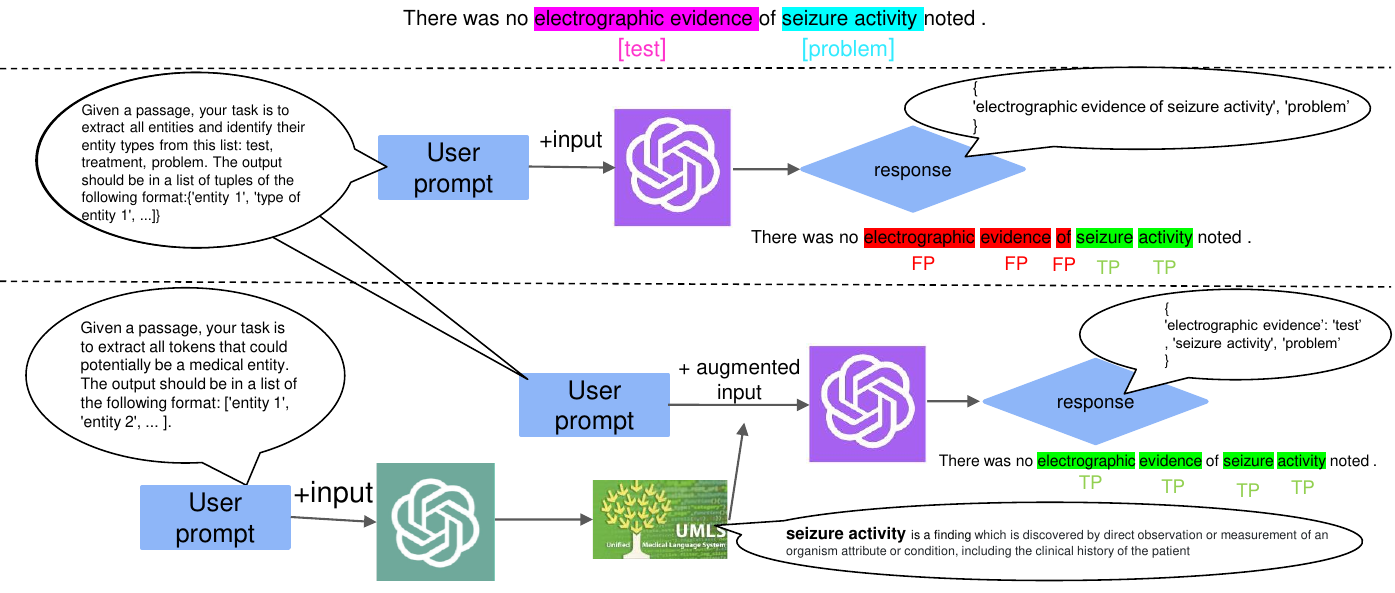}
    \put(0,39){\rotatebox{90}{\footnotesize{{input text}}}}
    \put(0,24){\rotatebox{90}{\footnotesize{{w/o DiRAG}}}}
    \put(0,9.5){\rotatebox{90}{\footnotesize{{w/ DiRAG}}}}
\end{overpic}
\caption{An overview of Dictionary-Infused RAG} \label{fig1}
\end{figure*}

\section{In-Context Learning or Fine-Tuning?}
Within the scope of LLMs for biomedical applications, an essential question is whether to prompt a closed-source LLM via ICL or fine-tune an open-source one. Comparing two different LLMs employing divergent strategies is not straightforward. To provide some insight into this dilemma, we examined two key factors, performance and cost, for biomedical NER, and present a detailed analysis under various experiment settings. This comparison offers valuable perspective into the right strategy given the task and dataset attributes. For fine-tuning, we used LoRA \cite{hu2021lora}. 
 Details can be found in Appendix \ref{sec:appendix}.
The cost of fine-tuning comes from training an LLM on a large labeled dataset while the cost of ICL mainly comes from calling an API for each input query. 
For 16-shot ICL experiments, we calculated the cost based on the number of processed and generated tokens considering the average text size based on current LLM API pricing.\footnote{ \url{https://openai.com/pricing}} The estimated cost for the entire test set of each benchmark dataset considering the input text, prompt, and generated text size using the TANL format is summarized in Table \ref{cost}. Referring to the OpenAI API for fine-tuning pricing, we also estimated the cost for fine-tuning LLama2-7B which is summarized in Table \ref{cost}. Interestingly, for the I2B2 dataset, GPT-3.5-turbo with a much cheaper cost outperforms fine-tuning Llama2-7B. 

\begin{table}
\small
\begin{center}
  \centering
  \resizebox{\linewidth}{!}{
  \begin{tabular}{ccccc}
    \toprule
    & Model & I2B2 & NCBI-disease &  BC2GM\\
    & & M/T & M/T &  M/T\\
    \midrule
    \multirow{4}{*}{Performance} & GPT-3.5-turbo w/ KATE  & 67.0/78.9 & 68.0/78.8 & 61.6/72.5 \\
    & GPT4 w/ KATE  & \textbf{81.7}/\textbf{88.4} &  79.3/88.3 & \textbf{72.4}/\textbf{80.7} \\
    & Llama2-7B  & 61.2/76.2 & \textbf{80.4}/\textbf{91.3} & 68.1/75.1  \\
    \midrule
    \multirow{4}{*}{Cost (T+I)} & GPT3.5-turbo w/ KATE  & (\$0.35) & (\$0.11) & (\$1.34) \\
    & GPT4 w/ KATE  & (\$10.42) & (\$3.12) &  (\$40.13) \\
    & Llama2-7B  & (\$47.85+\$7.4)  & (\$23.5+\$1.2) &  (\$69.7+\$12.9) \\

    \bottomrule
  \end{tabular}
  }
  \caption{Analysis of ICL vs fine-tuning LLMs: assessing performance and cost (Training + Inference) implications. Fine-tuning LLama2 exhibits superior outcomes on NCBI-disease, whereas GPT-4, enhanced by KATE using a biomedical encoder, achieves more favorable results on both the I2B2 and BC2GM datasets.
  }
  \label{cost}
  \end{center}
\vspace{-1pt}
\end{table}

\section{Dictionary-Infused RAG}
Retrieval-Augmented Generation (RAG) \cite{lewis2020retrieval} is a technique to enhance the capabilities of LLMs by integrating external information or knowledge into the generation process. This method involves retrieving relevant documents from a large corpus and providing this external knowledge in the input context to improve the quality and relevance of the generated text. 
Inspired by RAG, we developed a new method, DiRAG, to utilize UMLS as an external resource to augment the input data for the biomedical NER task. 
The process with detailed prompts is visualized in Fig \ref{fig1}, while an expanded view of the UMLS component is depicted in Fig \ref{umls2}. 
Unlike traditional RAG techniques that rely on embedding similarities to retrieve relevant documents, our approach initially employs the LLM to tackle a more straightforward task: identifying all words that could potentially qualify as medical named entities.
Then, we look up each selected word in an external knowledge base, e.g., UMLS to augment the input data with useful information such as term definition.  Then, we call the LLM with augmented input text. The process is visualized in Fig \ref{umls2}. 
We tested the approach on zero-shot NER and compared it with SOTA in Table \ref{RAG}. 
Our proposed approach enhanced the performance of both GPT versions on the I2B2 and NCBI-disease datasets significantly. DiRAG with GPT-4 achieved SOTA for zero-shot NER. Our approach proved ineffective for the BC2GM dataset due to the nature of the UMLS knowledge base which is predominantly tailored to medical terminology rather than biogenetics. We expect our approach to outperform GPT-4 on BC2GM with a more relevant knowledge base. 

\begin{table}
\small
\begin{center}
  \centering
  \resizebox{\linewidth}{!}{
  \begin{tabular}{cccc}
    \toprule
    Model & I2B2 & NCBI-disease &  BC2GM\\
    & M/T & M/T &  M/T\\
    \midrule
    UniversalNER \cite{zhou2023universalner} & 40.4/ - & 60.4/ - & 47.2/ - \\
    \cite{rohanian2023exploring} w/ GPT-3.5 & - & 
33.4 / - & 32.0 / - \\
    \cite{hu2024improving} w/ GPT-3.5-turbo & 39.3/ - & - & - \\
    \cite{hu2024improving} w/ GPT-4 & 52.6/ - & - & - \\
    \hdashline
    GPT-3.5-turbo w/o DiRAG & 41.9 /54.7 & 38.2 / 49.4  & 38.6 / 28.7 \\
    GPT-3.5-turbo w/ DiRAG & 43.0 / 55.7  & 44.7 / 50.0  & 30.45 / 22.5 \\
    \hdashline
    GPT-4 w/o DiRAG & 46.3 / 59.1 & 55.7 /60.5 &  \textbf{52.1} / \textbf{58.4}\\   
    GPT-4 w/ DiRAG  & \textbf{53.1 }/\textbf{62.8 } & \textbf{61.0 }/\textbf{66.2} & 51.1 / 55.0 \\

    \bottomrule
  \end{tabular}
  }
  \caption{Zero-shot NER with GPT models w/ and w/o DiRAG vs. SOTA. DiRAG improved zero-shot NER significantly for I2B2 and NCBI-disease datasets for both GPT models. Results with confidence intervals are in the appendix.}
  \label{RAG}
  \end{center}
\vspace{-1pt}
\end{table}

\section{Conclusion}
We explored LLMs for biomedical NER by customizing various prompting techniques.
Through a detailed comparative analysis, we highlighted the vital role of ICL and the selection of contextually pertinent examples with biomedical text encoders for biomedical NER tasks. Moreover, our investigation into incorporating external medical knowledge resulted in a novel data augmentation approach, considerably advancing the capabilities of zero-shot biomedical NER with LLMs. 

\section*{Limitations}
While we have shown the potential of enhancing LLM performance for biomedical NER, the experiments in this paper are limited in two aspects mainly due to computational constraints. (1) TANL uses a straightforward text-to-text format while DICE uses additional descriptions. Future work could attempt to simplify DICE or combine it with TANL.  Ablation studies on components of each format could help researchers design new prompt formatting strategies. (2) Our RAG-based method exclusively utilizes UMLS as the knowledge base, though it is limited in its vocabulary. For medical terms not covered by UMLS, we did not augment the input text. Other knowledge bases such as Wikipedia could serve as an alternative.




\nocite{Ando2005,andrew2007scalable,rasooli-tetrault-2015}

\bibliography{anthology,custom}

\begin{thebibliography}{55}
\expandafter\ifx\csname natexlab\endcsname\relax\def\natexlab#1{#1}\fi

\bibitem[{Alsentzer et~al.(2019)Alsentzer, Murphy, Boag, Weng, Jin, Naumann, and McDermott}]{alsentzer2019publicly}
Emily Alsentzer, John~R Murphy, Willie Boag, Wei-Hung Weng, Di~Jin, Tristan Naumann, and Matthew McDermott. 2019.
\newblock Publicly available clinical bert embeddings.
\newblock \emph{arXiv preprint arXiv:1904.03323}.

\bibitem[{Ando and Zhang(2005)}]{Ando2005}
Rie~Kubota Ando and Tong Zhang. 2005.
\newblock A framework for learning predictive structures from multiple tasks and unlabeled data.
\newblock \emph{Journal of Machine Learning Research}, 6:1817--1853.

\bibitem[{Andrew and Gao(2007)}]{andrew2007scalable}
Galen Andrew and Jianfeng Gao. 2007.
\newblock Scalable training of {L1}-regularized log-linear models.
\newblock In \emph{Proceedings of the 24th International Conference on Machine Learning}, pages 33--40.

\bibitem[{Bodenreider(2004)}]{bodenreider2004unified}
Olivier Bodenreider. 2004.
\newblock The unified medical language system (umls): integrating biomedical terminology.
\newblock \emph{Nucleic acids research}, 32(suppl\_1):D267--D270.

\bibitem[{Brown et~al.(2020)Brown, Mann, Ryder, Subbiah, Kaplan, Dhariwal, Neelakantan, Shyam, Sastry, Askell et~al.}]{brown2020language}
Tom Brown, Benjamin Mann, Nick Ryder, Melanie Subbiah, Jared~D Kaplan, Prafulla Dhariwal, Arvind Neelakantan, Pranav Shyam, Girish Sastry, Amanda Askell, et~al. 2020.
\newblock Language models are few-shot learners.
\newblock \emph{Advances in neural information processing systems}, 33:1877--1901.

\bibitem[{Chen et~al.(2023)Chen, Jiang, Chang, Hsieh, Yu, and Wang}]{chen2023minprompt}
Xiusi Chen, Jyun-Yu Jiang, Wei-Cheng Chang, Cho-Jui Hsieh, Hsiang-Fu Yu, and Wei Wang. 2023.
\newblock Minprompt: Graph-based minimal prompt data augmentation for few-shot question answering.
\newblock \emph{arXiv preprint arXiv:2310.05007}.

\bibitem[{Dai et~al.(2023)Dai, Sun, Dong, Hao, Ma, Sui, and Wei}]{dai2023can}
Damai Dai, Yutao Sun, Li~Dong, Yaru Hao, Shuming Ma, Zhifang Sui, and Furu Wei. 2023.
\newblock Why can gpt learn in-context? language models secretly perform gradient descent as meta-optimizers.
\newblock In \emph{Findings of the Association for Computational Linguistics: ACL 2023}, pages 4005--4019.

\bibitem[{Ding et~al.(2021)Ding, Hu, Zhao, Chen, Liu, Zheng, and Sun}]{ding2021openprompt}
Ning Ding, Shengding Hu, Weilin Zhao, Yulin Chen, Zhiyuan Liu, Hai-Tao Zheng, and Maosong Sun. 2021.
\newblock Openprompt: An open-source framework for prompt-learning.
\newblock \emph{arXiv preprint arXiv:2111.01998}.

\bibitem[{Do{\u{g}}an et~al.(2014)Do{\u{g}}an, Leaman, and Lu}]{dougan2014ncbi}
Rezarta~Islamaj Do{\u{g}}an, Robert Leaman, and Zhiyong Lu. 2014.
\newblock Ncbi disease corpus: a resource for disease name recognition and concept normalization.
\newblock \emph{Journal of biomedical informatics}, 47:1--10.

\bibitem[{Dong et~al.(2022)Dong, Li, Dai, Zheng, Wu, Chang, Sun, Xu, and Sui}]{dong2022survey}
Qingxiu Dong, Lei Li, Damai Dai, Ce~Zheng, Zhiyong Wu, Baobao Chang, Xu~Sun, Jingjing Xu, and Zhifang Sui. 2022.
\newblock A survey for in-context learning.
\newblock \emph{arXiv preprint arXiv:2301.00234}.

\bibitem[{Espejel et~al.(2023)Espejel, Ettifouri, Alassan, Chouham, and Dahhane}]{espejel2023gpt}
Jessica~L{\'o}pez Espejel, El~Hassane Ettifouri, Mahaman Sanoussi~Yahaya Alassan, El~Mehdi Chouham, and Walid Dahhane. 2023.
\newblock Gpt-3.5, gpt-4, or bard? evaluating llms reasoning ability in zero-shot setting and performance boosting through prompts.
\newblock \emph{Natural Language Processing Journal}, 5:100032.

\bibitem[{Gao et~al.(2023{\natexlab{a}})Gao, Wen, Gao, Wang, Zhang, and Lyu}]{gao2023makes}
Shuzheng Gao, Xin-Cheng Wen, Cuiyun Gao, Wenxuan Wang, Hongyu Zhang, and Michael~R Lyu. 2023{\natexlab{a}}.
\newblock What makes good in-context demonstrations for code intelligence tasks with llms?
\newblock In \emph{2023 38th IEEE/ACM International Conference on Automated Software Engineering (ASE)}, pages 761--773. IEEE.

\bibitem[{Gao et~al.(2021)Gao, Yao, and Chen}]{gao2021simcse}
Tianyu Gao, Xingcheng Yao, and Danqi Chen. 2021.
\newblock Simcse: Simple contrastive learning of sentence embeddings.
\newblock \emph{arXiv preprint arXiv:2104.08821}.

\bibitem[{Gao et~al.(2023{\natexlab{b}})Gao, Li, Caskey, Dligach, Miller, Churpek, and Afshar}]{gao2023leveraging}
Yanjun Gao, Ruizhe Li, John Caskey, Dmitriy Dligach, Timothy Miller, Matthew~M Churpek, and Majid Afshar. 2023{\natexlab{b}}.
\newblock Leveraging a medical knowledge graph into large language models for diagnosis prediction.
\newblock \emph{arXiv preprint arXiv:2308.14321}.

\bibitem[{Gao et~al.(2023{\natexlab{c}})Gao, Xiong, Gao, Jia, Pan, Bi, Dai, Sun, and Wang}]{gao2023retrieval}
Yunfan Gao, Yun Xiong, Xinyu Gao, Kangxiang Jia, Jinliu Pan, Yuxi Bi, Yi~Dai, Jiawei Sun, and Haofen Wang. 2023{\natexlab{c}}.
\newblock Retrieval-augmented generation for large language models: A survey.
\newblock \emph{arXiv preprint arXiv:2312.10997}.

\bibitem[{Grossman~Liu et~al.(2021)Grossman~Liu, Grossman, Mitchell, Weng, Natarajan, Hripcsak, and Vawdrey}]{grossman2021deep}
Lisa Grossman~Liu, Raymond~H Grossman, Elliot~G Mitchell, Chunhua Weng, Karthik Natarajan, George Hripcsak, and David~K Vawdrey. 2021.
\newblock A deep database of medical abbreviations and acronyms for natural language processing.
\newblock \emph{Scientific Data}, 8(1):149.

\bibitem[{Gu et~al.(2023)Gu, Zhang, Usuyama, Woldesenbet, Wong, Sanapathi, Wei, Valluri, Strandberg, Naumann et~al.}]{gu2023distilling}
Yu~Gu, Sheng Zhang, Naoto Usuyama, Yonas Woldesenbet, Cliff Wong, Praneeth Sanapathi, Mu~Wei, Naveen Valluri, Erika Strandberg, Tristan Naumann, et~al. 2023.
\newblock Distilling large language models for biomedical knowledge extraction: A case study on adverse drug events.
\newblock \emph{arXiv preprint arXiv:2307.06439}.

\bibitem[{Gururangan et~al.(2020)Gururangan, Marasović, Swayamdipta, Lo, Beltagy, Downey, and Smith}]{domains}
Suchin Gururangan, Ana Marasović, Swabha Swayamdipta, Kyle Lo, Iz~Beltagy, Doug Downey, and Noah~A. Smith. 2020.
\newblock Don't stop pretraining: Adapt language models to domains and tasks.
\newblock In \emph{Proceedings of ACL}.

\bibitem[{Gutierrez et~al.(2022)Gutierrez, McNeal, Washington, Chen, Li, Sun, and Su}]{gutierrez2022thinking}
Bernal~Jimenez Gutierrez, Nikolas McNeal, Clay Washington, You Chen, Lang Li, Huan Sun, and Yu~Su. 2022.
\newblock Thinking about gpt-3 in-context learning for biomedical ie? think again.
\newblock \emph{arXiv preprint arXiv:2203.08410}.

\bibitem[{Hsu et~al.(2021)Hsu, Huang, Boschee, Miller, Natarajan, Chang, Peng et~al.}]{hsu2021degree}
I~Hsu, Kuan-Hao Huang, Elizabeth Boschee, Scott Miller, Prem Natarajan, Kai-Wei Chang, Nanyun Peng, et~al. 2021.
\newblock Degree: A data-efficient generation-based event extraction model.
\newblock \emph{arXiv preprint arXiv:2108.12724}.

\bibitem[{Hu et~al.(2024{\natexlab{a}})Hu, Liu, Zhu, Lu, and Wu}]{hu2024zero}
Danqing Hu, Bing Liu, Xiaofeng Zhu, Xudong Lu, and Nan Wu. 2024{\natexlab{a}}.
\newblock Zero-shot information extraction from radiological reports using chatgpt.
\newblock \emph{International Journal of Medical Informatics}, 183:105321.

\bibitem[{Hu et~al.(2021)Hu, Shen, Wallis, Allen-Zhu, Li, Wang, Wang, and Chen}]{hu2021lora}
Edward~J Hu, Yelong Shen, Phillip Wallis, Zeyuan Allen-Zhu, Yuanzhi Li, Shean Wang, Lu~Wang, and Weizhu Chen. 2021.
\newblock Lora: Low-rank adaptation of large language models.
\newblock \emph{arXiv preprint arXiv:2106.09685}.

\bibitem[{Hu et~al.(2024{\natexlab{b}})Hu, Chen, Du, Peng, Keloth, Zuo, Zhou, Li, Jiang, Lu et~al.}]{hu2024improving}
Yan Hu, Qingyu Chen, Jingcheng Du, Xueqing Peng, Vipina~Kuttichi Keloth, Xu~Zuo, Yujia Zhou, Zehan Li, Xiaoqian Jiang, Zhiyong Lu, et~al. 2024{\natexlab{b}}.
\newblock Improving large language models for clinical named entity recognition via prompt engineering.
\newblock \emph{Journal of the American Medical Informatics Association}, page ocad259.

\bibitem[{Kaddour et~al.(2023)Kaddour, Harris, Mozes, Bradley, Raileanu, and McHardy}]{kaddour2023challenges}
Jean Kaddour, Joshua Harris, Maximilian Mozes, Herbie Bradley, Roberta Raileanu, and Robert McHardy. 2023.
\newblock Challenges and applications of large language models.
\newblock \emph{arXiv preprint arXiv:2307.10169}.

\bibitem[{Karabacak and Margetis(2023)}]{karabacak2023embracing}
Mert Karabacak and Konstantinos Margetis. 2023.
\newblock Embracing large language models for medical applications: Opportunities and challenges.
\newblock \emph{Cureus}, 15(5).

\bibitem[{Kumari et~al.(2023)Kumari, Kumari, Singh, Singh, Juhi, Dhanvijay, Pinjar, Mondal, and Dhanvijay}]{kumari2023large}
Amita Kumari, Anita Kumari, Amita Singh, Sanjeet~K Singh, Ayesha Juhi, Anup Kumar~D Dhanvijay, Mohammed~Jaffer Pinjar, Himel Mondal, and Anoop~Kumar Dhanvijay. 2023.
\newblock Large language models in hematology case solving: a comparative study of chatgpt-3.5, google bard, and microsoft bing.
\newblock \emph{Cureus}, 15(8).

\bibitem[{Lehman et~al.(2023)Lehman, Hernandez, Mahajan, Wulff, Smith, Ziegler, Nadler, Szolovits, Johnson, and Alsentzer}]{lehman2023we}
Eric Lehman, Evan Hernandez, Diwakar Mahajan, Jonas Wulff, Micah~J Smith, Zachary Ziegler, Daniel Nadler, Peter Szolovits, Alistair Johnson, and Emily Alsentzer. 2023.
\newblock Do we still need clinical language models?
\newblock \emph{arXiv preprint arXiv:2302.08091}.

\bibitem[{Lester et~al.(2021)Lester, Al-Rfou, and Constant}]{lester2021power}
Brian Lester, Rami Al-Rfou, and Noah Constant. 2021.
\newblock The power of scale for parameter-efficient prompt tuning.
\newblock \emph{arXiv preprint arXiv:2104.08691}.

\bibitem[{Lewis et~al.(2020)Lewis, Perez, Piktus, Petroni, Karpukhin, Goyal, K{\"u}ttler, Lewis, Yih, Rockt{\"a}schel et~al.}]{lewis2020retrieval}
Patrick Lewis, Ethan Perez, Aleksandra Piktus, Fabio Petroni, Vladimir Karpukhin, Naman Goyal, Heinrich K{\"u}ttler, Mike Lewis, Wen-tau Yih, Tim Rockt{\"a}schel, et~al. 2020.
\newblock Retrieval-augmented generation for knowledge-intensive nlp tasks.
\newblock \emph{Advances in Neural Information Processing Systems}, 33:9459--9474.

\bibitem[{Liu et~al.(2021)Liu, Shen, Zhang, Dolan, Carin, and Chen}]{liu2021makes}
Jiachang Liu, Dinghan Shen, Yizhe Zhang, Bill Dolan, Lawrence Carin, and Weizhu Chen. 2021.
\newblock What makes good in-context examples for gpt-$3 $?
\newblock \emph{arXiv preprint arXiv:2101.06804}.

\bibitem[{Lu et~al.(2021)Lu, Bartolo, Moore, Riedel, and Stenetorp}]{lu2021fantastically}
Yao Lu, Max Bartolo, Alastair Moore, Sebastian Riedel, and Pontus Stenetorp. 2021.
\newblock Fantastically ordered prompts and where to find them: Overcoming few-shot prompt order sensitivity.
\newblock \emph{arXiv preprint arXiv:2104.08786}.

\bibitem[{Ma et~al.(2022)Ma, Taylor, Wang, and Peng}]{ma2022dice}
Mingyu~Derek Ma, Alexander~K Taylor, Wei Wang, and Nanyun Peng. 2022.
\newblock Dice: data-efficient clinical event extraction with generative models.
\newblock \emph{arXiv preprint arXiv:2208.07989}.

\bibitem[{Min et~al.(2022)Min, Lyu, Holtzman, Artetxe, Lewis, Hajishirzi, and Zettlemoyer}]{min2022rethinking}
Sewon Min, Xinxi Lyu, Ari Holtzman, Mikel Artetxe, Mike Lewis, Hannaneh Hajishirzi, and Luke Zettlemoyer. 2022.
\newblock Rethinking the role of demonstrations: What makes in-context learning work?
\newblock \emph{arXiv preprint arXiv:2202.12837}.

\bibitem[{Moradi et~al.(2021)Moradi, Blagec, Haberl, and Samwald}]{moradi2021gpt}
Milad Moradi, Kathrin Blagec, Florian Haberl, and Matthias Samwald. 2021.
\newblock Gpt-3 models are poor few-shot learners in the biomedical domain.
\newblock \emph{arXiv preprint arXiv:2109.02555}.

\bibitem[{Nori et~al.(2023{\natexlab{a}})Nori, King, McKinney, Carignan, and Horvitz}]{nori2023capabilities}
Harsha Nori, Nicholas King, Scott~Mayer McKinney, Dean Carignan, and Eric Horvitz. 2023{\natexlab{a}}.
\newblock Capabilities of gpt-4 on medical challenge problems.
\newblock \emph{arXiv preprint arXiv:2303.13375}.

\bibitem[{Nori et~al.(2023{\natexlab{b}})Nori, Lee, Zhang, Carignan, Edgar, Fusi, King, Larson, Li, Liu et~al.}]{nori2023can}
Harsha Nori, Yin~Tat Lee, Sheng Zhang, Dean Carignan, Richard Edgar, Nicolo Fusi, Nicholas King, Jonathan Larson, Yuanzhi Li, Weishung Liu, et~al. 2023{\natexlab{b}}.
\newblock Can generalist foundation models outcompete special-purpose tuning? case study in medicine.
\newblock \emph{arXiv preprint arXiv:2311.16452}.

\bibitem[{Paolini et~al.(2021)Paolini, Athiwaratkun, Krone, Ma, Achille, Anubhai, Santos, Xiang, and Soatto}]{paolini2021structured}
Giovanni Paolini, Ben Athiwaratkun, Jason Krone, Jie Ma, Alessandro Achille, Rishita Anubhai, Cicero Nogueira~dos Santos, Bing Xiang, and Stefano Soatto. 2021.
\newblock Structured prediction as translation between augmented natural languages.
\newblock \emph{arXiv preprint arXiv:2101.05779}.

\bibitem[{Ramshaw and Marcus(1999)}]{ramshaw1999text}
Lance~A Ramshaw and Mitchell~P Marcus. 1999.
\newblock Text chunking using transformation-based learning.
\newblock In \emph{Natural language processing using very large corpora}, pages 157--176. Springer.

\bibitem[{Rasooli and Tetreault(2015)}]{rasooli-tetrault-2015}
Mohammad~Sadegh Rasooli and Joel~R. Tetreault. 2015.
\newblock \href {http://arxiv.org/abs/1503.06733} {Yara parser: {A} fast and accurate dependency parser}.
\newblock \emph{Computing Research Repository}, arXiv:1503.06733.
\newblock Version 2.

\bibitem[{Reimers and Gurevych(2019)}]{reimers-2019-sentence-bert}
Nils Reimers and Iryna Gurevych. 2019.
\newblock \href {https://arxiv.org/abs/1908.10084} {Sentence-bert: Sentence embeddings using siamese bert-networks}.
\newblock In \emph{Proceedings of the 2019 Conference on Empirical Methods in Natural Language Processing}. Association for Computational Linguistics.

\bibitem[{Rohanian et~al.(2023)Rohanian, Nouriborji, and Clifton}]{rohanian2023exploring}
Omid Rohanian, Mohammadmahdi Nouriborji, and David~A Clifton. 2023.
\newblock Exploring the effectiveness of instruction tuning in biomedical language processing.
\newblock \emph{arXiv preprint arXiv:2401.00579}.

\bibitem[{Seo et~al.(2024)Seo, Baek, Thorne, and Hwang}]{seo2024retrieval}
Minju Seo, Jinheon Baek, James Thorne, and Sung~Ju Hwang. 2024.
\newblock Retrieval-augmented data augmentation for low-resource domain tasks.
\newblock \emph{arXiv preprint arXiv:2402.13482}.

\bibitem[{Smith et~al.(2008)Smith, Tanabe, Ando, Kuo, Chung, Hsu, Lin, Klinger, Friedrich, Ganchev et~al.}]{smith2008overview}
Larry Smith, Lorraine~K Tanabe, Rie Johnson~nee Ando, Cheng-Ju Kuo, I-Fang Chung, Chun-Nan Hsu, Yu-Shi Lin, Roman Klinger, Christoph~M Friedrich, Kuzman Ganchev, et~al. 2008.
\newblock Overview of biocreative ii gene mention recognition.
\newblock \emph{Genome biology}, 9:1--19.

\bibitem[{Song et~al.(2020)Song, Tan, Qin, Lu, and Liu}]{song2020mpnet}
Kaitao Song, Xu~Tan, Tao Qin, Jianfeng Lu, and Tie-Yan Liu. 2020.
\newblock Mpnet: Masked and permuted pre-training for language understanding.
\newblock \emph{Advances in Neural Information Processing Systems}, 33:16857--16867.

\bibitem[{Tian et~al.(2024)Tian, Jin, Yeganova, Lai, Zhu, Chen, Yang, Chen, Kim, Comeau et~al.}]{tian2024opportunities}
Shubo Tian, Qiao Jin, Lana Yeganova, Po-Ting Lai, Qingqing Zhu, Xiuying Chen, Yifan Yang, Qingyu Chen, Won Kim, Donald~C Comeau, et~al. 2024.
\newblock Opportunities and challenges for chatgpt and large language models in biomedicine and health.
\newblock \emph{Briefings in Bioinformatics}, 25(1):bbad493.

\bibitem[{Uzuner et~al.(2011)Uzuner, South, Shen, and DuVall}]{uzuner20112010}
{\"O}zlem Uzuner, Brett~R South, Shuying Shen, and Scott~L DuVall. 2011.
\newblock 2010 i2b2/va challenge on concepts, assertions, and relations in clinical text.
\newblock \emph{Journal of the American Medical Informatics Association}, 18(5):552--556.

\bibitem[{Vaishya et~al.(2023)Vaishya, Misra, and Vaish}]{vaishya2023chatgpt}
Raju Vaishya, Anoop Misra, and Abhishek Vaish. 2023.
\newblock Chatgpt: Is this version good for healthcare and research?
\newblock \emph{Diabetes \& Metabolic Syndrome: Clinical Research \& Reviews}, 17(4):102744.

\bibitem[{Wang et~al.(2023{\natexlab{a}})Wang, Zhao, Ouyang, Wang, and Shen}]{wang2023chatcad}
Sheng Wang, Zihao Zhao, Xi~Ouyang, Qian Wang, and Dinggang Shen. 2023{\natexlab{a}}.
\newblock Chatcad: Interactive computer-aided diagnosis on medical image using large language models.
\newblock \emph{arXiv preprint arXiv:2302.07257}.

\bibitem[{Wang et~al.(2023{\natexlab{b}})Wang, Sun, Li, Ouyang, Wu, Zhang, Li, and Wang}]{wang2023gpt}
Shuhe Wang, Xiaofei Sun, Xiaoya Li, Rongbin Ouyang, Fei Wu, Tianwei Zhang, Jiwei Li, and Guoyin Wang. 2023{\natexlab{b}}.
\newblock Gpt-ner: Named entity recognition via large language models.
\newblock \emph{arXiv preprint arXiv:2304.10428}.

\bibitem[{Wang et~al.(2023{\natexlab{c}})Wang, Zhou, Zu, Xia, Chen, Zhang, Zheng, Ye, Zhang, Gui et~al.}]{wang2023instructuie}
Xiao Wang, Weikang Zhou, Can Zu, Han Xia, Tianze Chen, Yuansen Zhang, Rui Zheng, Junjie Ye, Qi~Zhang, Tao Gui, et~al. 2023{\natexlab{c}}.
\newblock Instructuie: Multi-task instruction tuning for unified information extraction.
\newblock \emph{arXiv preprint arXiv:2304.08085}.

\bibitem[{Weber et~al.(2021)Weber, S{\"a}nger, M{\"u}nchmeyer, Habibi, Leser, and Akbik}]{weber2021hunflair}
Leon Weber, Mario S{\"a}nger, Jannes M{\"u}nchmeyer, Maryam Habibi, Ulf Leser, and Alan Akbik. 2021.
\newblock Hunflair: an easy-to-use tool for state-of-the-art biomedical named entity recognition.
\newblock \emph{Bioinformatics}, 37(17):2792--2794.

\bibitem[{Webson and Pavlick(2021)}]{webson2021prompt}
Albert Webson and Ellie Pavlick. 2021.
\newblock Do prompt-based models really understand the meaning of their prompts?
\newblock \emph{arXiv preprint arXiv:2109.01247}.

\bibitem[{White et~al.(2023)White, Fu, Hays, Sandborn, Olea, Gilbert, Elnashar, Spencer-Smith, and Schmidt}]{white2023prompt}
Jules White, Quchen Fu, Sam Hays, Michael Sandborn, Carlos Olea, Henry Gilbert, Ashraf Elnashar, Jesse Spencer-Smith, and Douglas~C Schmidt. 2023.
\newblock A prompt pattern catalog to enhance prompt engineering with chatgpt.
\newblock \emph{arXiv preprint arXiv:2302.11382}.

\bibitem[{Zakka et~al.(2024)Zakka, Shad, Chaurasia, Dalal, Kim, Moor, Fong, Phillips, Alexander, Ashley et~al.}]{zakka2024almanac}
Cyril Zakka, Rohan Shad, Akash Chaurasia, Alex~R Dalal, Jennifer~L Kim, Michael Moor, Robyn Fong, Curran Phillips, Kevin Alexander, Euan Ashley, et~al. 2024.
\newblock Almanac—retrieval-augmented language models for clinical medicine.
\newblock \emph{NEJM AI}, 1(2):AIoa2300068.

\bibitem[{Zhou et~al.(2023)Zhou, Zhang, Gu, Chen, and Poon}]{zhou2023universalner}
Wenxuan Zhou, Sheng Zhang, Yu~Gu, Muhao Chen, and Hoifung Poon. 2023.
\newblock Universalner: Targeted distillation from large language models for open named entity recognition.
\newblock \emph{arXiv preprint arXiv:2308.03279}.

\end{thebibliography}

\appendix

\section{Appendix}

\subsection{TANL/DICE more examples}
In Fig 4-8, we visualize two examples of each format for NCBI-disease and BC2GM datasets for more demonstration.

\subsection{Benchmark datasets}
We studied LLMs on I2b2, NCBI-disease, and BC2GM dataset. In the following, we provide some details about each.

\textbf{I2B2}: I2B2 is a collection of annotated clinical records that are used primarily for Clinical NER.
The task involves identifying clinical terms such as medical problems, treatments, and tests from patient records. The dataset typically includes a large number of annotated clinical narratives that are de-identified to protect patient confidentiality. This makes it a rich resource for training and testing NER models.

\textbf{NCBI-disease}: This dataset is specifically curated for disease name recognition and normalization in biomedical texts. It comprises abstracts from PubMed annotated for disease mentions and linked to the NCBI disease database. The corpus is relatively smaller compared to i2b2 but is densely annotated, providing high-quality, fine-grained annotations of disease entities, which are crucial for models aimed at medical literature.

\textbf{BC2GM}: This dataset focuses on the recognition of gene and gene product mentions in PubMed abstracts that is a suitable dataset for biological NER. The BC2GM dataset is extensively annotated to include a wide range of gene and gene product mentions, reflecting the complex and varied ways these entities are referred to in scientific literature.

\subsection{PEFT setting of Llama for fine-tuning} We fine-tuned Llama2-7B on the entire training set of each dataset for three epochs and maintained a batch size of 16, learning rate of 2e-4, and cap the maximum sequence length at 512, truncating any sequences that exceeded this limit. The LoRA dropout rate is adjusted to 0.1, and the LoRA $\alpha$ and rank parameters are also set at 16 and 32 respectively. The training was done on 4 NVIDIA Tesla V100 GPUs for approximately 24, 12, and 63 hours for I2B2, NCBI-disease, and BC2GM respectively.

\subsection{Few-shot and Zero-shot performances with Confidence Interval} \label{app_p3}
We introduced both few-shot and zero-shot settings to comprehensively evaluate the versatility and generalization capabilities of our study across different levels of data availability. While it's true that the performance in the zero-shot setting is generally lower compared to the few-shot setting, this approach offers valuable insights into the model's behavior when no training examples are provided. The zero-shot setting, leveraging techniques like Retrieval-Augmented Generation (RAG), demonstrates the model's potential to utilize pre-existing knowledge embedded in its parameters and external sources effectively. This is particularly important for scenarios where labeled data is scarce or unavailable, making zero-shot learning a critical area of study to ensure broader applicability of the model in real-world applications. Moreover, the inclusion of both methodologies allows us to highlight the performance trade-offs and strengths of the model under different instructional paradigms, contributing to a more robust and nuanced understanding of its capabilities. We ran all experiments with different random seeds and reported the full results of Table \ref{KATE}, \ref{cost}, and \ref{RAG} with confidence Intervals In Tables \ref{apx1}, \ref{apx2}, and \ref{apx3}.

\subsection{UMLS detail} \label{app_p4}
In Fig \ref{umls2}, we visualize the process by which potential words suggested by the LLM are searched within the UMLS and demonstrate how the input is augmented to enhance zero-shot prompting in LLMs.

\label{sec:appendix}


\begin{figure*}[ht]
\centering
\begin{overpic}
[width=0.8\textwidth]{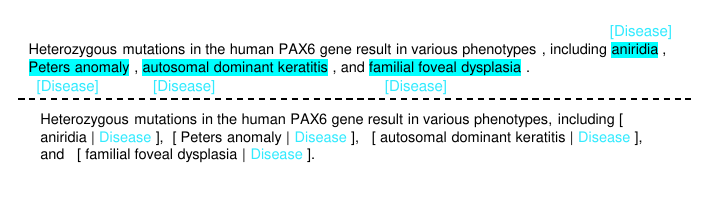}
    \put(0,16){\rotatebox{90}{\footnotesize{{input}}}}
    \put(0,6){\rotatebox{90}{\footnotesize{{output}}}}
\end{overpic}
\caption{TANL input-output format example for NCBI-disease dataset} \label{tanl_disease}
\end{figure*}

\begin{figure*}[ht]
\centering
\begin{overpic}
[width=0.8\textwidth]{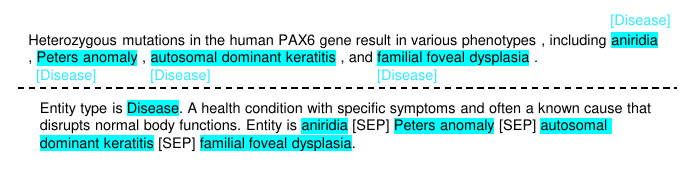}
    \put(0,15){\rotatebox{90}{\footnotesize{{input}}}}
    \put(0,5){\rotatebox{90}{\footnotesize{{output}}}}
\end{overpic}
\caption{DICE input-output format example for NCBI-disease dataset} \label{dice_disease}
\end{figure*}

\begin{figure*}[ht]
\centering
\begin{overpic}
[width=0.8\textwidth]{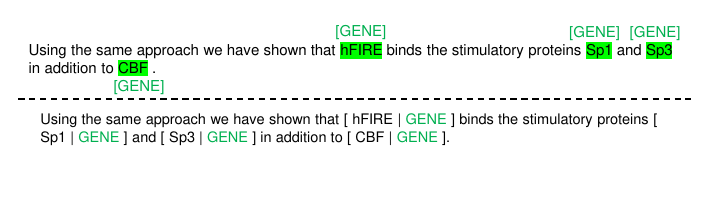}
    \put(0,16.5){\rotatebox{90}{\footnotesize{{input}}}}
    \put(0,6.5){\rotatebox{90}{\footnotesize{{output}}}}
\end{overpic}
\caption{TANL input-output format example for BC2GM dataset} \label{tanl_gene}
\end{figure*}

\begin{figure*}[ht]
\centering
\begin{overpic}
[width=0.8\textwidth]{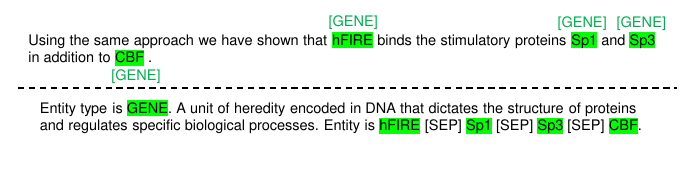}
    \put(0,16){\rotatebox{90}{\footnotesize{{input}}}}
    \put(0,5.5){\rotatebox{90}{\footnotesize{{output}}}}
\end{overpic}
\caption{DICE input-output format example for BC2GM dataset} \label{dice-gene}
\end{figure*}

\begin{table*}
\small
\begin{center}
  \centering
  \resizebox{\linewidth}{!}{
  \begin{tabular}{ccccc}
    \toprule
    & & & Mention/Token &  \\
    Model & input-output format & I2B2 & NCBI-disease &  BC2GM\\
    \midrule
    \multirow{2}{*}{GPT-3.5-turbo}
    & DICE  & 41.2 $\pm$ 0.2 /50.0 $\pm$ 0.1 & 45.3 $\pm$ 0.2 /\textbf{62.0 $\pm$ 0.3} & \textbf{43.3 $\pm$ 0.5} /\textbf{55.6 $\pm$ 0.4}  \\
    & TANL  & \textbf{52.9 $\pm$ 0.3}/\textbf{59.7 $\pm$ 0.4} & \textbf{46.5 $\pm$ 0.5}/51.3 $\pm$ 0.5 & 39.1 $\pm$ 0.4/50.8 $\pm$ 0.5 \\
    
    \midrule
    \multirow{2}{*}{GPT-4} 
    & DICE  & 58.8 $\pm$ 0.4 /70.1 $\pm$ 0.3 & \textbf{68.1 $\pm$ 0.9}/\textbf{77.8 $\pm$ 1.1} &  \textbf{57.1 $\pm$ 0.6}/67.9 $\pm$ 0.5 \\
    & TANL  & \textbf{61.9 $\pm$ 0.3}/\textbf{73.5 $\pm$ 0.5} &  67.5 $\pm$ 0.8/70.0 $\pm$ 0.6 & 56.4 $\pm$ 0.2/\textbf{69.6 $\pm$ 0.3}  \\
    
    \bottomrule
  \end{tabular}
  }
  \caption{TANL vs. DICE format with GPT-3.5-turbo/GPT-4 with confidence intervals} 
  \label{dicevsTANL2}
  \end{center}
\vspace{-1pt}
\end{table*}

\begin{table*}[ht]
\small
\begin{center}
  \centering
  \resizebox{2\columnwidth}{!}{
  \begin{tabular}{ccccc}
    \toprule
    Model & KATE vs RS & I2B2 & NCBI-disease &  BC2GM \\
    & & M/T & M/T &  M/T\\
    \midrule
    \multirow{6}{*}{GPT-3.5-turbo (ICL)} 
    & RS & 52.9 $\pm$ 0.3 / 59.7 $\pm$ 0.4 & 46.6 $\pm$ 0.5 / 51.3 $\pm$ 0.5 & 39.1 $\pm$ 0.4 / 50.8 $\pm$ 0.5 \\
    \cdashline{2-5}
    & BioClinicalRoBERTa & 66.1 $\pm$ 0.4 / 77.4 $\pm$ 0.6 & \textbf{68.0 $\pm$ 0.3 }/ 77.7 $\pm$ 0.2 & \textbf{61.6 $\pm$ 0.5} / \textbf{72.5 $\pm$ 0.6} \\
    & BioClinicalBERT  & \textbf{67.0 $\pm$ 0.6}/\textbf{78.9 $\pm$ 0.5}  & 67.6 $\pm$ 0.1/\textbf{78.8$\pm$ 0.1} & 60.9 $\pm$ 0.7 /72.0 $\pm$ 0.5  \\
    \cdashline{2-5}
    & MPNET  & 65.3 $\pm$ 0.3 / 76.7 $\pm$ 0.2 & 63.7 $\pm$ 0.3 / 76.7 $\pm$ 0.3 &  59.1 $\pm$ 0.4 / 70.0 $\pm$ 0.4 \\
    & SimCSE  & 65.2 $\pm$ 0.2 / 76.1 $\pm$ 0.3  & 61.6 $\pm$ 0.4 / 76.1 $\pm$ 0.3 & 57.8 $\pm$ 0.5 / 68.8 $\pm$ 0.4  \\
    \cline{2-5}
    & \cite{hu2024improving} & 49.3/ - & - & - \\
    \midrule
    \multirow{6}{*}{GPT4 (ICL)} 
    & RS & 67.7 $\pm$ 0.3 / 73.5 $\pm$ 0.5 & 62.6 $\pm$ 0.8 /70.0 $\pm$ 0.6 & 59.2 $\pm$ 0.2 / 69.6 $\pm$ 0.3 \\
    \cdashline{2-5}
    & BioClinicalRoBERTa & 81.2 $\pm$ 0.3/\textbf{88.4 $\pm$ 0.6} & \textbf{79.3 $\pm$ 0.9}/\textbf{88.3 $\pm$ 0.8} & \textbf{72.4 $\pm$ 0.6}/\textbf{80.7 $\pm$ 0.5} \\
    & BioClinicalBERT  & \textbf{81.7 $\pm$ 0.4}/88.1 $\pm$ 0.3  & \textbf{79.3 $\pm$ 0.4}/88.0 $\pm$ 0.3 & 71.9 $\pm$ 0.3 / 79.4 $\pm$ 0.3 \\
    \cdashline{2-5}
    & MPNET & 80.7 $\pm$ 0.4 / 87.5 $\pm$ 0.5 & 79.8 $\pm$ 0.9 /87.4 $\pm$ 0.9 &  71.1 $\pm$ 1.1 / 80.2 $\pm$ 1.0 \\
    & SimCSE  & 79.6 $\pm$ 0.5 / 86.6 $\pm$ 0.4  & 77.3 $\pm$ 0.5 / 86.5 $\pm$ 0.8 & 69.9 $\pm$ 0.8 / 77.9 $\pm$ 0.5  \\

    \cline{2-5}
    & \cite{hu2024improving} & 59.3/ - & - & - \\
    
    \midrule
    BioBERT & fully supervised & - /87.3 & - /\textbf{89.1} &  - /83.8 \\
    BioClinicBERT & fully supervised & - /87.7 & - /89.0 &  - /81.7 \\
    BioClinicRoBERTa & fully supervised & - /\textbf{89.7} & - /89.0 &  - /\textbf{87.0} \\
    
    \bottomrule
  \end{tabular}
  }
  \caption{Random example selection (RS) vs. KATE with medical/non-medical encoders vs. fully supervised models with Mention/Token-level (M/T) analysis. KATE significantly outperforms random sampling in all settings, and LMs pre-trained on the biomedical text outperform strong, general domain encoders. HunFlair is added to the paper}
  \label{apx1}
  \end{center}
\vspace{-1pt}
\end{table*}

\begin{table*}
\small
\begin{center}
  \centering
  \resizebox{\linewidth}{!}{
  \begin{tabular}{ccccc}
    \toprule
    & Model & I2B2 & NCBI-disease &  BC2GM\\
    & & M/T & M/T &  M/T\\
    \midrule
    \multirow{4}{*}{Performance} & GPT-3.5-turbo w/ KATE  & 67.0 $\pm$ 0.6 / 78.9 $\pm$ 0.5 & 68.0 $\pm$ 0.3 / 78.8 $\pm$ 0.1 & 61.6 $\pm$ 0.1 / 72.5 $\pm$ 0.6 \\
    & GPT4 w/ KATE  & \textbf{81.7 $\pm$ 0.4} / \textbf{88.4 $\pm$ 0.6} &  79.3 $\pm$ 0.4 / 88.3 $\pm$ 0.8 & \textbf{72.4 $\pm$ 0.6} / \textbf{80.7 $\pm$ 0.4} \\
    & Llama2-7B  & 61.2 $\pm$ 1.8 / 76.2 $\pm$ 1.3 & \textbf{80.4 $\pm$ 0.9}/ \textbf{91.3 $\pm$ 1.1} & 68.1 $\pm$ 1.4 / 75.1 $\pm$ 1.3  \\
    \midrule
    \multirow{4}{*}{Cost (T+I)} & GPT3.5-turbo w/ KATE  & (\$0.35) & (\$0.11) & (\$1.34) \\
    & GPT4 w/ KATE  & (\$10.42) & (\$3.12) &  (\$40.13) \\
    & Llama2-7B  & (\$47.85+\$7.4)  & (\$23.5+\$1.2) &  (\$69.7+\$12.9) \\

    \bottomrule
  \end{tabular}
  }
  \caption{Analysis of ICL vs fine-tuning LLMs: assessing performance and cost implications. Fine-tuning LLama2 exhibits superior outcomes on NCBI-disease, whereas GPT-4, enhanced by KATE using a biomedical encoder, achieves more favorable results on both the I2B2 and BC2GM datasets.
  }
  \label{apx2}
  \end{center}
\vspace{-1pt}
\end{table*}

\begin{table*}
\small
\begin{center}
  \centering
  \resizebox{2\columnwidth}{!}{
  \begin{tabular}{cccc}
    \toprule
    Model & I2B2 & NCBI-disease &  BC2GM\\
    & M/T & M/T &  M/T\\
    \midrule
    UniversalNER \cite{zhou2023universalner} & 40.4/ - & 60.4/ - & 47.2/ - \\
    \cite{rohanian2023exploring} w/ GPT-3.5 & - & 
33.4 / - & 32.0 / - \\
    \cite{hu2024improving} w/ GPT-3.5-turbo & 39.3/ - & - & - \\
    \cite{hu2024improving} w/ GPT-4 & 52.6/ - & - & - \\
    HunFlair \cite{weber2021hunflair} & 0.0 / 0.0 & 24.8 / 36.1 & 28.2 / 22.7  \\
    \hdashline
    GPT-3.5-turbo w/o DiRAG & 41.9 $\pm$ 1.4 /54.7 $\pm$ 1.9 & 38.2 $\pm$ 1.7 / 49.4 $\pm$ 2.6 & 38.6 $\pm$ 1.0 / 28.7 $\pm$ 1.9 \\
    GPT-3.5-turbo w/ DiRAG & 43.0 $\pm$ 0.9 / 55.7 $\pm$ 1.5 & 44.7 $\pm$ 0.5 / 50.0 $\pm$ 2.1 & 30.45 $\pm$ 1.6 / 22.5 $\pm$ 2.1 \\
    \hdashline
    GPT-4 w/o DiRAG & 46.3 $\pm$ 1.9 / 59.1 $\pm$ 2.7 & 55.7 $\pm$ 0.8 /60.5 $\pm$ 0.9 &  \textbf{52.1 $\pm$ 3.64} / \textbf{58.4 $\pm$ 1.3}\\   
    GPT-4 w/ DiRAG  & \textbf{53.1 $\pm$ 1.1 }/\textbf{	62.8 $\pm$ 1.2 } & \textbf{61.0 $\pm$ 0.6 }/\textbf{66.2 $\pm$ 0.5} & 51.1 $\pm$ 2.0 / 55.0 $\pm$ 2.2 \\

    \bottomrule
  \end{tabular}
  }
  \caption{Full results of Zero-shot NER with GPT-3.5-turbo and GPT-4, w/ and w/o DiRAG, and their comparision with SOTA. Our method improved zero-shot NER significantly for I2B2 and NCBI-disease datasets.}
  \label{apx3}
  \end{center}
\vspace{-1pt}
\end{table*}

\begin{figure*}[ht]
\centering
\begin{overpic}
[width=0.8\textwidth]{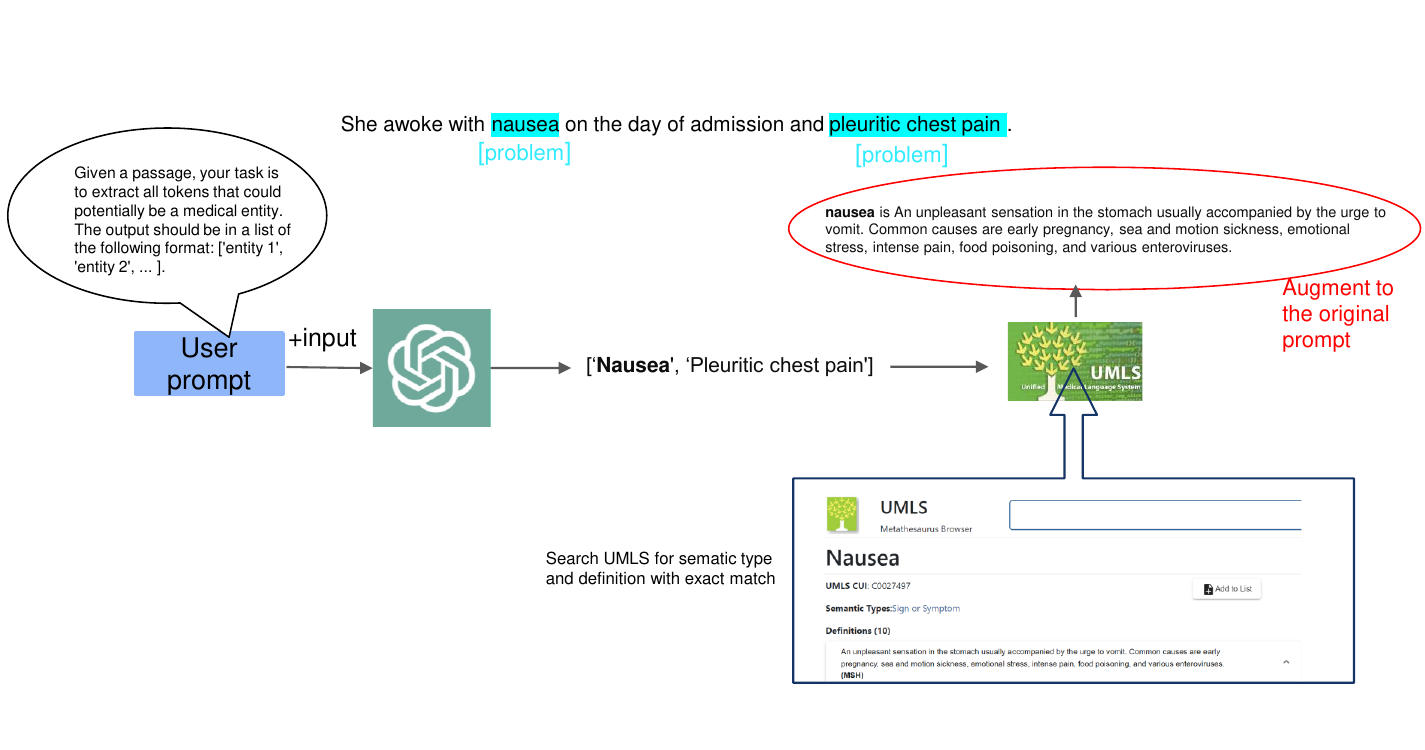}
\end{overpic}
\caption{UMLS search. The GPT model is prompted for a simpler task of identifying all words that could potentially be a named entity. Then, the retrieved information from UMLS will augment the original input text for recalling the LLM} \label{umls2}
\end{figure*}

\end{document}